\documentclass{article}

\usepackage{microtype}
\usepackage{graphicx}
\usepackage{wrapfig}
\usepackage{subfigure}
\usepackage{booktabs} % for professional tables

\usepackage{hyperref}

\usepackage{amssymb} 
\usepackage{amsmath}
\usepackage{amsthm}
\usepackage{bbm}
\usepackage{dsfont}
\usepackage{graphicx}
\usepackage{color}
\usepackage{subfigure}
\usepackage{float}
\usepackage{xfrac}
\usepackage{natbib}
\graphicspath{{../figs/},{../GenExamples/}}
\definecolor{Red}{rgb}{1.0,0,0} 
\definecolor{Red}{rgb}{1.0,0,0} 

\newcommand{\comm}[1]{\textcolor{Red}{[#1]}} % --- comment out when DONE with revisions ---
\newcommand{\pmm}{$PM_{2.5}\,$}
\usepackage{xcolor}
\usepackage{xspace}
\makeatletter
\DeclareRobustCommand\onedot{\futurelet\@let@token\@onedot}
\def\@onedot{\ifx\@let@token.\else.\null\fi\xspace}

\def\eg{{e.g}\onedot} \def\Eg{{E.g}\onedot}

\makeatother
\usepackage{xargs}                      % Use more than one optional parameter in a new commands
\usepackage[colorinlistoftodos,prependcaption,textsize=tiny]{todonotes}
\newcommandx{\unsure}[2][1=]{\todo[linecolor=red,backgroundcolor=red!25,bordercolor=red,#1]{#2}}
\newcommandx{\change}[2][1=]{\todo[linecolor=blue,backgroundcolor=blue!25,bordercolor=blue,#1]{#2}}
\newcommandx{\info}[2][1=]{\todo[linecolor=OliveGreen,backgroundcolor=OliveGreen!25,bordercolor=OliveGreen,#1]{#2}}
\newcommandx{\improvement}[2][1=]{\todo[linecolor=Plum,backgroundcolor=Plum!25,bordercolor=Plum,#1]{#2}}
\newcommandx{\thiswillnotshow}[2][1=]{\todo[disable,#1]{#2}}
\newcommand\todocomment[1]{\textcolor{red}{#1}}

\usepackage[accepted]{icml2020}
\icmltitlerunning{Predicting High-risk SARS-CoV-2 Regions with U-Net Driven Quantile Regression} % TODO 
\begin{document}

\twocolumn[
\icmltitle{In the Danger Zone: U-Net Driven Quantile Regression can Predict High-risk SARS-CoV-2 
            Regions via Pollutant Particulate Matter and Satellite Imagery
%\icmltitle{In the Danger Zone: Learning High-risk SARS-CoV-2 Regions with Quantile \\
 %           Regression Unets using Pollutant particulate Matter and Satellite Imagery
           }  



\icmlsetsymbol{equal}{*}

%XXX
\begin{icmlauthorlist}
\icmlauthor{Jacquelyn A. Shelton}{polyu}
\icmlauthor{Przemyslaw Polewski}{tt,polyu}
% \icmlauthor{Drew Lipman}{dre} # XXX FIXME change if drew gets the go ahead
\icmlauthor{Wei Yao}{polyu}
\end{icmlauthorlist}

\icmlaffiliation{polyu}{Department of Land Surveying and Geoinformatics,
The Hong Kong Polytechnic University, Hong Kong SAR, China}
\icmlaffiliation{tt}{TomTom Location Technology Germany GmbH, Berlin, Germany}
%\icmlaffiliation{dre}{XXX, Washington DC, USA}

\icmlcorrespondingauthor{Jacquelyn A. Shelton}{jacquelyn.ann.shelton@gmail.com}
%\icmlcorrespondingauthor{Eee Pppp}{ep@eden.co.uk}

\icmlkeywords{COVID19, SARS-CoV-2, Pollutant Particulate Matter, Satellite Imagery, U-net, Remote Sensing, Convolutional Neural Networks}

\vskip 0.3in
]



\printAffiliationsAndNotice{}  % leave blank if no need to mention equal contribution
%\printAffiliationsAndNotice{\icmlEqualContribution} % otherwise use the
% standard text.

% TODO
\begin{abstract}
Since the outbreak of COVID-19 policy makers have been relying upon non-pharmacological interventions to control the outbreak.
With air pollution as a potential transmission vector there is need to include it in intervention strategies.
We propose a U-net driven quantile regression model to predict $PM_{2.5}$ air pollution based on easily obtainable satellite imagery.
We demonstrate that our approach can reconstruct \pmm concentrations on ground-truth data and predict reasonable \pmm values with
their spatial distribution, even for locations where pollution data is unavailable.
Such predictions of \pmm characteristics could crucially advise public policy strategies geared to reduce the transmission of and lethality of COVID-19.

\end{abstract}

\section{Introduction}
\label{sec:intro}
\vspace{-.1cm}

% XXX3
% \comm{Use $\backslash comm \{text\}$ to make any comments/changes} % TODO comment \comm command out in preamble above
% XXX

%The method can learn a model of how the pollution particulates them self actually look, so that in an unseen state it can see similar type data and tell you where there's pollutant particulates

%thus this is a question of critical importance.

Since the outbreak of the Severe Acute Respiratory Syndrome Corona Virus 2 (SARS-CoV-2), popularly referred to as \emph{COVID-19}, many questions have been asked around the disease.
Of particular interest are the questions of transmission methods, and characteristics that identify vulnerable populations.
Studies into these properties are made more complex by the lack of information
around the behavior of the virus, \Eg the percent of cases that
are asymptomatic.
However, as indicated by the inclusion in the CDC's \href{https://www.cdc.gov/coronavirus/2019-ncov/need-extra-precautions/people-at-higher-risk.html}{ vulnerable Population index}, asthma and chronic lung disease seem to be a major factor in the severity of the cases.
For example, in \cite{pmid32268945} the connections between the lethality rate in Lombardy and Emilia Romagna, areas with high level of atmospheric pollution, are explored.
In particular, the paper studies the correlation between pollution, which is a known instigator of chronic lung disease even in young and other wise healthy subjects, and the lethality of SARS-CoV-2.
A similar result for the lethality in the United States
\cite{Wu2020.04.05.20054502} using county level fatality rate, and county level
long term air pollution, shows that, after adjusting for other known factors,
that there is a strong correlation between the concentration of particulate
matter $2.5$ micrometers or less in diameter, or $PM_{2.5}$, and the county level
lethality. Specifically, a 1 $\frac{\mu g}{m^3}$ increase in $PM_{2.5}$
corresponds to an $8\%$ increase in the fatality rate.

%\change{remove specific references to pm10 and replace with particulate
%matter, since we are not dealing with pm10 here}

However, as observed in \cite{Coccia2020.04.06.20055657}, there is also a correlation between particulate matter ($PM$) air pollution and the number of reported cases.
This suggests that pollution-to-human may serve as another transmission dynamic for SARS-CoV-2.
These two results suggest a two factor vulnerability to SARS-CoV-2 caused by increased particulate matter in the air:
On one hand it increases the likelihood of having a more sever reaction to infection, and the other it serves a transmission vector.
\begin{figure*}[t]
\centering
\includegraphics[width=.91\linewidth]{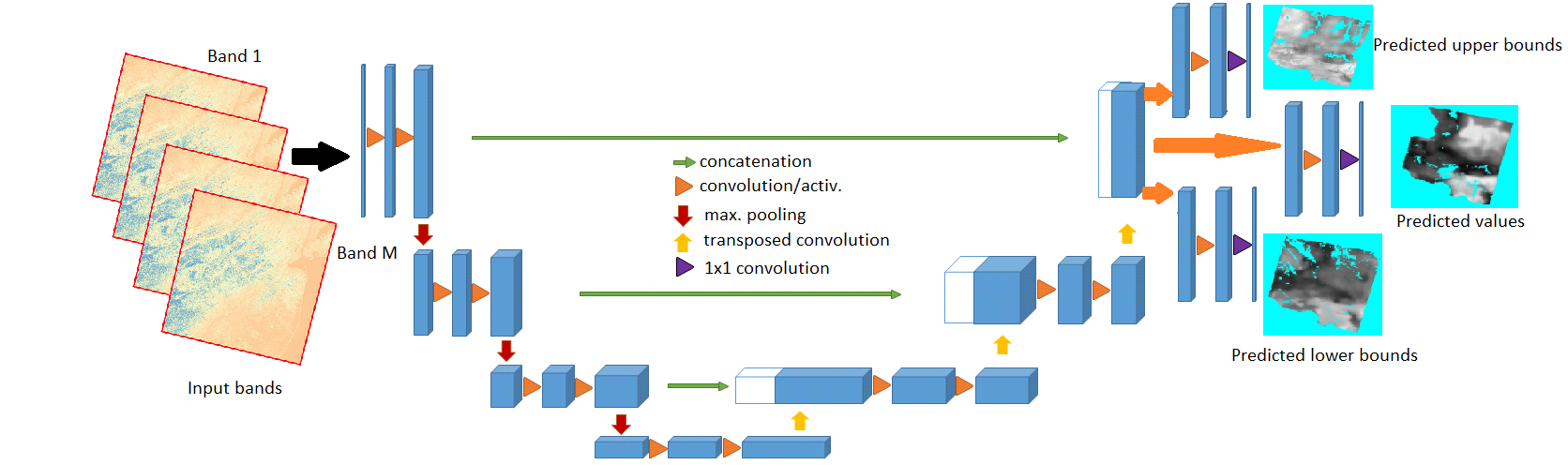}
\vspace{-.5cm}
\caption{Network structure for predicting \pmm concentrations from
multispectral imagery, based on the classic
U-Net~\citep{unet_paper} architecture up to the top-level upsampling layer,
where 3 parallel sequences of convolutional filters and activations
branch off the upsampled features, corresponding to the 3 types of outputs
predicted by the network (lower/upper bound, median).}
\label{fig:unet}
\end{figure*}

Traditionally,  $PM_{2.5}$ concentration data can be obtained from ground sensors and
measurement stations. However, the spatial resolution of these measurements is
greatly limited by the sparsity of sensor networks. 
Thus, detailed pollution maps have been developed that integrate
heterogeneous data sources (see e.g.~\citet{pm_chemcomposition2019}),
%proposed a $PM_{2.5}$ estimation of fusing aerosol optical depthinformation computed from satellite imagery with a chemical transport model and
%regional ground-based observations.
%They published 
to create a database of estimated monthly pollutant concentrations over several countries.
This data was used by % as the basis for the aforementioned study by
\citet{Wu2020.04.05.20054502} linking $PM_{2.5}$ concentrations to COVID-19
mortality rates.
Although this historical data helped
establish this causality relationship, up-to-date/live pollution information is 
still needed to monitor pollutant $PM_{2.5}$.

% XXX HAD TO CUT SORRY XXX TODO FIXME
%Recently, there has been a growing interest in retrieving spatio-temporal air
%quality products based on high resolution optical satellite missions such as
%\textit{Landsat 8}~\citep{LandsatRef}, utilizing top-of-atmosphere image reflectances as
%a basis for directly predicting pollution
%parameters~\citep{YANG2020140,doi:10.1029/2018JD028759}. 
% cut?

%The primary goal of this work is to provide a machine learning model capable of estimating the $PM_{2.5}$ concentration with comparable granularity to the data made available by \citet{pm_chemcomposition2019}, based on publicly available satellite imagery. This could provide researchers and decision makers with a tool to estimate dynamic particulate matter concentrations at a monthly scale, to facilitate further COVID-19 studies and to support strategic planning of infection management and response.

% XXX NEW: GOOD POINT FROM REV 1 XXX
By knowing where pollution is we can better understand the impacts of social segregation policy on subpopulations in order to e.g. allocate medical funds to the most vulnerable populaces.
The health needs for a population often require very granular data about their circumstances, and air pollution is one such measure that is only sporadically known in poorer areas and thus is insufficiently represented in modeling the health situation.
In order to properly control COVID-19 hotspots there is a need to predict the spread and intensity of COVID-19. 
While contact tracing and non-pharmacological interventions (NPIs) have shown value, there is mounting evidence that there are other transmission vectors, including via pollution particles, as discussed earlier. 
Therefore there is mounting need to understand local pollution dynamics in order to correctly deal with the pandemic. 
In order to properly understand the effectiveness of a given NPI it is essential we understand the main transmission vectors. It is, therefore, vital we gain understanding of the strength of air pollution as a transmission vector in order to understand and design NPIs efficiently.
% XXX

The goal of the present work is to model air pollution \pmm concentrations using readily available satellite imagery.
The aim is to aid in any planning for efficacious COVID-19 geared strategies. 
The paper is organized as follows: 
Sec.~\ref{sec:model} introduces the proposed U-net model, Sec.~\ref{sec:data} presents the data, 
Sec.~\ref{sec:exps} describes the experiments and results, and finally Sec.~\ref{sec:disc} provides a summary and outlook.

\vspace{-.2cm}

\section{U-net model for pollutant particulate matter}
\label{sec:model}
\vspace{-.1cm}
We build upon the well known U-net architecture~\cite{unet_paper} to predict
dense (per-pixel) $PM_{2.5}$ concentrations from multispectral satellite
imagery. Since its introduction in 2015, the U-net has been successfully
applied to various semantic segmentation tasks, \Eg in medical
imaging~\citep{10.1007/978-3-319-60964-5_44} and astronomy\citep{akeret2017radio}.
The U-net consists of two symmetrical parts. The encoder path downsamples the
original image into meaningful features by means of convolutional filters and
pooling operations, whereas the upsampling path aims at decoding these features
into a full-sized output map using transposed convolution operations, driven by
an appropriate loss function. Moreover, upsampling layers are augmented
with feature maps from the downsampling path at the corresponding resolution,
to provide more context information.
The original U-net was meant for classification and featured a softmax layer after the top-level upsampling
layer's output, trained using a cross-entropy objective with discrete
ground-truth labels. Yet the U-net has also been used for dense regression by
removing the softmax layer and optimizing the squared difference between the
upsampled output and a continuous target
variable~\citep{YAO2018364,10.1007/978-3-319-46723-8_27}.

It is well known that least squares regression estimates the conditional mean of
the response variable given the predictors, and is therefore sensitive to
outliers. Alternatively, Quantile regression~\citep{koenker_2005} is more robust to outliers.
Let $S=\{(x_i,y_i)\}, 1 \leq i \leq N$ denote
a set of predictor variable vectors $x$ and matching continuous response
variables $y$ sampled from their respective distributions $\mathcal{X},\mathcal{Y}$.
Let $f(x|\theta)$ indicate the prediction function parameterized by $\theta$, and let $0 < q < 1$ refer to the
quantile level. In this work, we consider losses of the form:
\vspace{-.3cm}
\begin{equation}\label{eq:quantileLoss}
\small
\begin{split}
L_q(\theta) &= \frac{1}{|S|}\sum_{(x,y) \in S}\rho_q(y - f(x|\theta))\\
\rho_q(r) &= r[r \geq 0] - (1-q)r
\end{split}
\end{equation}
\normalsize
\vspace{-.1cm}
In the above, $[a]$ denotes the indicator function for event $a$, whereas the
term $\rho_q(r)$ is the \emph{check function}~\citep{koenker_2005}. To better
quantify the approximated distribution, we learn 3 quantiles $q_l <
q_{m}=0.5 < q_u$ simultaneously, which allows us to obtain both a point-wise
estimate from the median $q_m$ and a $(q_u - q_l)-$confidence interval for the
value. We cast the aggregate loss function as a linear combination of partial
$L_q$ terms with coefficients $\gamma_l, \gamma_u$, in a simple multi-task
learning setting~\citep{8848395}:
\vspace{-.2cm}
\begin{equation}\label{eq:aggrLoss}
L_{aggr}(\theta) = \gamma_l L_{q_l}(\theta) + L_{0.5}(\theta) + \gamma_u
L_{q_u}(\theta)
\end{equation}
\vspace{-.1cm}
\normalsize
To mirror the structure of the loss function, the U-net architecture was
extended to include a separate sequence of convolutional filters at the top
level of the upsampling path per quantile loss term (see Fig.~\ref{fig:unet}).
One disadvantage of using the loss $L_q$ (Eq.~\ref{eq:quantileLoss}) is the fact
that the check function $\rho_q$ is not differentiable at zero. Moreover, the
derivative is piece-wise constant on $R^+,R^-$. This might pose a challenge to
gradient-based optimization schemes (\Eg back propagation in neural networks)
because the gradient norm does not get smaller as the optimization converges to
a local minimum. To alleviate that, some approximations of the check function
have been proposed. The Huber loss~\cite{hastie01statisticallearning}
combines quadratic behavior within a $\delta-$neighborhood of 0 with linear
behavior on $R \setminus [-\delta;\delta]$. This can be used to approximate
$\rho_{0.5}$.
Recently,~\citet{Gupta2020} introduced an asymmetric version of the Huber loss:
\vspace{-.3cm}
%\small
\begin{equation}
H(r|\delta_l,\delta_u) = r^2 - (r-\delta_l)_{+}^{2} - (-r-\delta_u)_{+}^{2}
\label{eq:ass-huber}
\end{equation}
%\normalsize
%\vspace{-.2cm}
%
\begin{figure}[t]
\begin{minipage}[b]{0.65\linewidth}
\centering
\includegraphics[width=1.0\linewidth]{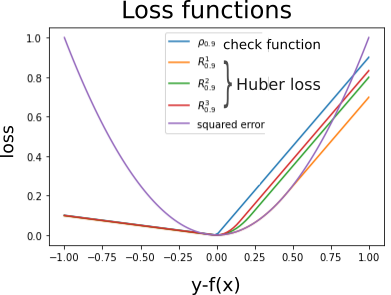}
\vspace{-1cm}
\end{minipage}
\begin{minipage}[b]{0.32\linewidth}
\caption{Comparison of loss functions: the differentiable asymmetric Huber
function in Eq.~\eqref{eq:ass-huber} for various check function $\rho$
approximations versus the standard squared loss.} \label{fig:losses}
\end{minipage}
%\vspace{-4mm}
\end{figure}
This loss function is differentiable everywhere. The parameters
$\delta_l,\delta_u$ control both the slope of the linear functions on their
respective sides of the real axis, and the locations where the function starts
to show quadratic behavior. We propose to approximate $\rho_q$ with
\small $R^\alpha_{q}(r)=\alpha H(r|q/2\alpha,(1-q)/2\alpha)$\normalsize, where
$\alpha$ is a parameter which controls the location of the change from linear to quadratic
characteristics. See Fig.~\ref{fig:losses} for a comparison of loss functions.

\vspace{-.2cm}

\section{Data}
\label{sec:data} 
\vspace{-.1cm}
%
%\comm{merge some of intro text here?}
%In order to address this problem we use sentinal satellite data~\cite{main_sat_source}, which is updated every 2 weeks, and ppm2.5 data 
%Maybe don't want citations here, getting redundant...
%
In order to address this problem, we use satellite imagery as a source
of predictor variables, and approximate $PM_{2.5}$ concentrations published
by~\citet{pm_chemcomposition2019} for the time period of
2000-2018 in the role of ground-truth. 

\textbf{Satellite data.}
%Sentinel images contain X spectral bands. We consider 10 of these: . These bands allow for the estimation of Aerosol Optical Depth, which represents the concentration of fine particulates in the air. 
%Preprocessing... removed cloud structure with mask to prevent interference
Our work used Landsat 8 satellite imagery published by the
United States Geological Survey~\cite{LandsatRef}. 
Landsat 8 is the latest 
in a series of Earth observation missions %dating back to 1972 (Landsat 1), 
containing a total of $11$ spectral bands, ranging in
wavelengths from $0.435 \mu m$ to $12.51 \mu m$, with spatial resolution between $15$ to
$100m$ depending on the band.

\textbf{Pollution data.} 
%how this pm2.5 map was created ~\cite{pm_chemcomposition2019}
%it has some really good info in the introduction\todocomment{fill in details}
We downloaded monthly $PM_{2.5}$ concentration maps for North America 
%the FTP server of Dalhousie University, Canada, published by Aaron van Donkelaar
%and collaborators
from~\citep{Donkelaar_PM2.5_FTP}. 
Readily available, these maps contain $0.01$ degrees per pixel and use a standard WGS84 coordinate
reference system. We used the data made available immediately after Landsat 8 mission launched, 
namely from March 2013 to December 2018.

\textbf{Preprocessing.}
The Landsat imagery was first reprojected to the WGS84 coordinate system, and
downsampled to the ground-truth resolution of $0.01$ degrees (the panchromatic
band was dropped). To match the temporal resolution of the ground-truth, we computed
per-band average images grouped by month of acquisition. Next, we
derived a per-pixel mask of regions within the image covered by cirrus clouds
and hence not suitable for analysis, based on the estimated
cloud cover percentage~\citep{abc2017}. Finally, we combined
the cloud cover mask with the data availability mask from the ground-truth
$PM_{2.5}$ maps. Also, pixels corresponding to the top and bottom $1\%$ of
ground-truth values were masked out as outliers. All input imagery bands were normalized to the
interval $[0;1]$ individually per band.

\vspace{-.2cm}
\section{Experiments}
\label{sec:exps}
\begin{figure}[ht]
%\vskip 0.2in
\begin{center}
\centerline{\includegraphics[width=1\columnwidth]{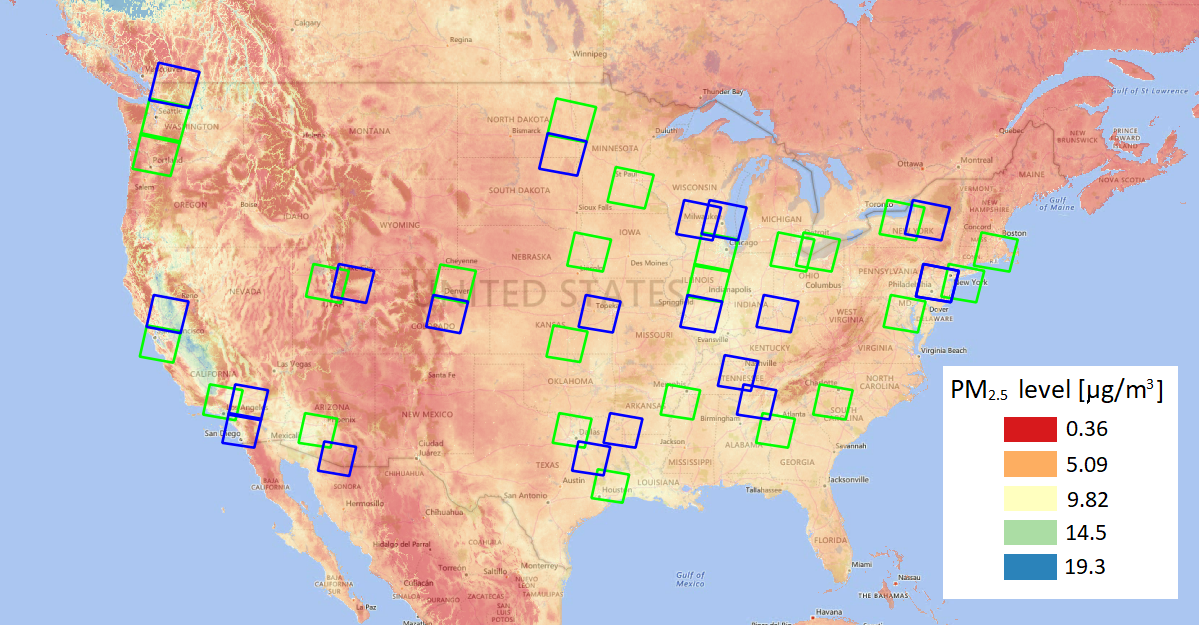}}
\vspace{-.3cm}
\caption{Locations of US cities chosen as a basis for our study. Rectangles
indicate bounding boxes of the downloaded Landsat images. Locations marked with
green boxes were used for training and validation, whereas blue boxes represent
regions used exclusively for testing. The background color map represents
average concentration of \pmm  in 2018.}
\label{fig:trainTest}
\end{center}
%\vskip -0.2in
\end{figure}

\vspace{-.1cm}
%
%
%The overarching goal of all experiments is to learn and infer the structure,
%spatial distribution of and physical values of pollutant particulate matter
%\pmm concentrations using satellite imagery paired with particulate matter
% data.
%
We selected an initial number $N = 133$ Landsat images, spanning March 2013 to
December 2018, of $24$ major cities from representative regions of
the United States (see Fig.~\ref{fig:trainTest}, green boxes) and preprocessed
them as described in Sec.~\ref{sec:data}.
Next, we defined our ground-truth data as these images paired with corresponding $PM_{2.5}$ images 
from which we use an 80:20$\%$ random split to create training/testing set of 106:27 images. 
%at that it had been trained on, but to time
%however at  on imagery from previously unseen points in time. 
%
A U-net based on adapting the implementation by~\citet{akeret2017radio} was
trained using an ADAM optimized over $1000$ epochs with minibatch size of $15$
and $100$ internal iterations,
%It was implemented in tensorflow based on github repo xyz adapted for xxx, 
with the following parameterization: dropout ratio was $0.5$, learning rate
$0.00005$, regression quantiles $q_l=0.1, q_r=0.9$.
For the Huber loss function~\eqref{eq:ass-huber}, all
aggregates~\eqref{eq:aggrLoss} contributed in equal proportion and the
parameter controlling the function shape was $\alpha = 2$.

\begin{figure}[t]
%\begin{minipage}[b]{0.6\linewidth}
\centering
\includegraphics[width=.85\linewidth]{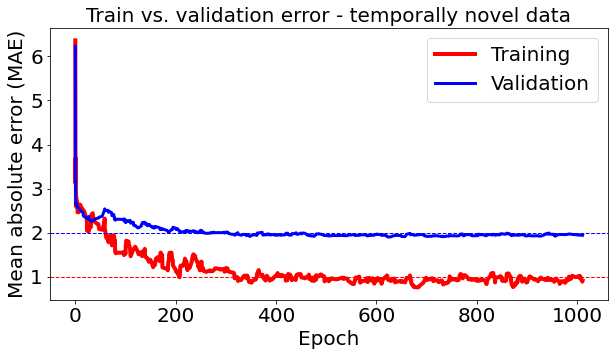}
\vspace{-.4cm}
%\end{minipage}
%\begin{minipage}[b]{0.38\linewidth}
\caption{%Error predicting \pmm concentrations with ground-truth data. 
The error of the U-net trained in Exp.~4.1 shown with the mean absolute error (MAE).
Red curve: error predicting \pmm concentrations of data with ground-truth. 
Blue curve: error predicting \pmm of validation data in Exp.~4.2, where the
U-net was generalized to cities at times not present in the training data but locations it has been trained on.
Error with the GT data converges to $MAE \approx 1$ and the error with validation data converges to $MAE \approx 2$. 
This shows that the U-net can generalize well to temporally novel data.
  } 
\label{fig:exp1-errors}
%\end{minipage}
%\vspace{-4mm}
\end{figure}

\begin{figure}[t]
\begin{minipage}[b]{0.56\linewidth}
\centering
\includegraphics[width=1.0\linewidth]{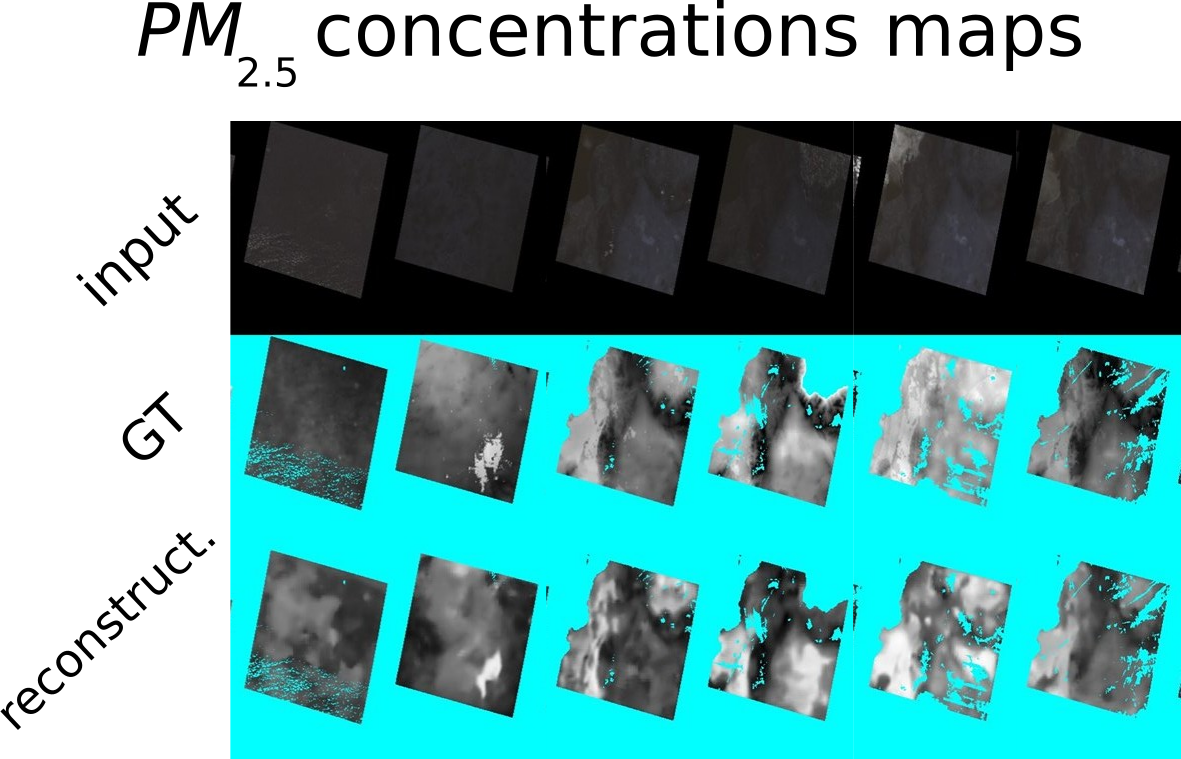}
%\vspace{-1cm}
\end{minipage}
%\vspace{-1.1cm}
\hspace{.04cm}
\begin{minipage}[b]{0.34\linewidth}
\caption{%Reconstruction of ground-truth \pmm pollution concentrations.
The U-net trained on ground-truth satellite images was able to successfully
reconstruct the \pmm concentrations, corresponding to the low error shown
in Fig.~\ref{fig:exp1-errors}. %Larger values indicate higher pollution. }
Shown here are a few examples.}
\label{fig:gt-reconstruct}
\end{minipage}
%\vspace{-4mm}
\end{figure}

\textbf{4.1: Sanity check with ground-truth.} % 1
The network converged to a state which produced a mean absolute error of
$\approx 1 \mu g/m^3$ between predicted and ground-truth values on the
training set (see Fig.~\ref{fig:exp1-errors}). The median width of the predicted
confidence intervals (upper bound - lower bound) for training data was $1.94 \mu
g/m^3$, and $70\%$ of ground-truth values fell into the predicted interval. The
number of ground-truth values above the lower bound and below the
upper bound was respectively $88\%$ and $82\%$. This demonstrates that the model
was able to approximate the $0.1$ and $0.9$ quantiles well, and also to provide
quality point-estimates.
Correspondingly, our approach can predict \pmm pollution maps that match
the structure and approximate concentrations of ground-truth reconstructions.
See Fig.~\ref{fig:gt-reconstruct} for some examples.

\textbf{4.2: Generalizability to temporally novel data.} % 2
To assess the ability of our model to generalize in the temporal domain, we
utilized $27$ previously unseen images that overlapped with the training set
spatially but not temporally. The test set contained images from $15$ of the $24$
cities. The mean absolute validation error followed the same trend as Exp.~4.1
with GT training data, however it converged to about $2 \mu g/m^3$
(see Fig.~\ref{fig:exp1-errors}). The performance of the predicted confidence
intervals degraded to 40\% of contained ground-truth values, whereas the median
interval width remained low at $2.06 \mu g/m^3$. This indicates a degree of
overfitting, however the predicted \pmm concentrations are still within
reasonable distance of the ground-truth.

\textbf{4.3: Cities with similar pollution profile.} % 3
We selected $28$ new images at $20$ additional locations within the
United States, showing visually similar distributions of ground-truth $PM_{2.5}$
within the Landsat image frames to the original training cities,
% TODO do we need to formalize this, like the 2-sample test or some histogram similarity metric
in order to evaluate the performance of predicting the \pmm concentrations at
locations unseen during training. The obtained mean absolute error was $2.81 \mu
g/m^3$, and the predicted confidence intervals contained $31\%$ of ground-truth
values. This shows that it is harder to generalize in the spatial than in the
temporal domain.

\begin{figure*}[t]
%\begin{minipage}[b]{0.6\linewidth}
\centering
\includegraphics[width=1\linewidth]{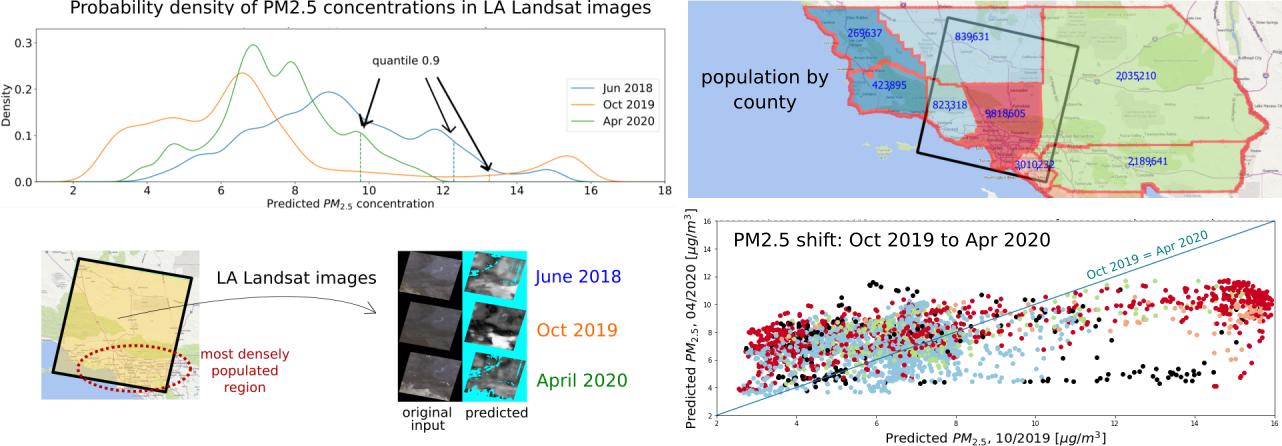}
%\vspace{-.8cm}
%\end{minipage}
%\begin{minipage}[b]{0.38\linewidth}
\caption{Density of predicted \pmm concentrations in Los Angeles for time
periods before and during the COVID-19 outbreak (top left), corresponding
Landsat images with their predicted \pmm pollution maps (bottom left).
As expected, we see a significant reduction in the $0.9$-quantile of the predicted \pmm pollution distributions 
between April 2020 (immediately after the lock-down began) and the previous years. 
The predicted \pmm maps match this trend, showing relatively less pollution in 2020 (darker pixels). 
Interestingly, yet unsurprisingly, the most densely populated counties, Los Angeles and
Orange County, (encircled in red on the map) are predicted to have higher \pmm
concentrations for all of the years, as shown in the \pmm pollution maps. 
The 2010 population map shows this dense region had a joint population of $12.8$ million (top right), 
and between October 2019 and April 2020, this region showed the greatest shift in predicted \pmm concentrations (bottom right).
The scatter plot of a sample of predictions from October 2019 versus April 2020 (teal line represents
identity mapping) from 2010 (bottom right) illustrates this \pmm shift, where points are color-coded by originating county (black points
correspond to the ocean).
The most significant drop in predicted \pmm concentration occurs in Los Angeles (red points) and
Orange County (peach).}
\label{fig:LA-prob-dens}
%\end{minipage}
\end{figure*}

\textbf{4.4: Before and after SARS-CoV-2-induced lock-down.} % 5
The goal of the fourth experiment is to verify that our model can predict expected \pmm concentration trends at time points before and after the government mandated 
lock-down in March 2020 intended to hinder the spread of SARS-CoV-2 (which consequently \eg drastically reduced industrial emissions). 
We applied the U-net learned in Exp.~4.3 to satellite images of Los Angeles, CA from 2018, 2019, and early 2020. 
The results, shown in Fig.~\ref{fig:LA-prob-dens}, illustrate that the U-net was able to learn \pmm concentrations consistent with world events at the time. 
Namely, the pollution is inferred to be notably higher before the lock-down than after the lock-down, shown by the considerable shift of the predicted \pmm distribution's $0.9$ quantile between October 2019 and April 2020 -- the quantile shifted from $13.2$ to $9.7$, respectively.
Additionally, our approach was able to successfully isolate regions in LA that are the most densely populated by predicting higher pollutant \pmm values. 
%\todocomm{the U-net was successfully able to learn the characteristics of the particulate matter with an \comm{error of} and \comm{x mass within y quantile}, following trends consistent with current events.}
%%See Fig.~\ref{fig:LA-prob-dens} for illustrations with the plotted probability density estimates the predicted \pmm concentrations for each time point and the corresponding reconstructions of the \pmm images.
This suggests that our approach can generalize to new data and reliably predict pollutant \pmm concentrations and their spatial structure, which informs on the presence and lethality of SARS-CoV-2.

\iffalse 
%TODO FIXME INCLUDE IF ACTUALLY RUN EXPERIMENT AND HAVE SOMETHING TO SEE
\textbf{4.5: Generalizability to never before seen locations.} % 4
Finally, we test the model of Exp.~4.3 on cities for which \textit{no pollutant \pmm data} 
is available from any time.
\comm{ Although we of course have no ground-truth \pmm data with which to compare these results, we can 
rely on these results to be within blah standard deviations from the true \pmm values and that they 
can isolate the structure of the pollutants across a given city, because the \pmm reconstruction performance reflects that of the learning of similar cities where we do have ground-truth
it would need to at least semi reliably work in areas where we know nothing about the pollution
so sure, it's not going to be amazing and we can't verify it
but it's the best anyone can do currently, its definitely better than nothing
and it should be at least as good as our predictions on unseen cities AS LONG as the cities are "similar"...a topic to quantify in future work.}
\fi % XXX XXX 

%\comm{See full set of results in Supplementary Material A. (tables etc, graphs
% for all exps)}

\iffalse
\textbf{Evaluation metrics.}
\todocomment{what eval metrics were used? rmse, mean absolute dev, quantiles,
number of GT values within the predicted confidence interval, per training,
'same city different time' testing, 'different city' testing? figure of abs dev
per test image?}

\textbf{Results.}
\todocomment{what to present?? table with metric names on one axis and data set
(training, test tier 1, test tier 2?), plots of neural network loss or
validation metrics per epoch? images of reconstructed \pmm concentration?}
\fi

\vspace{-.2cm} 

\section{Discussion}
\label{sec:disc}
\vspace{-.1cm}
SARS-CoV-2 lethality has been concretely linked to the concentration of pollutant particulate matter, where a slight increase in \pmm can drastically enhance the morbidity rate. 
We have proposed a means of learning the structure, spatial distribution, and physical concentration of pollutant particulate matter \pmm based solely on 
global satellite imagery and monthly \pmm values from the years 2013-2018. 
Our approach proposes a U-net convolutional network with quantile regression loss to learn dense per-pixel \pmm prediction maps.
We have demonstrated not only that our approach can successfully reconstruct \pmm concentrations with low error, but it can also generalize to completely unseen locations for which pollutant data is not available. 
%This enables the identification of dangerous regions where SARS-CoV-2 is particularly transmissable and lethal.
%Important to know where morbibity is likely to be high such that e.g. intervention policies can be implemented correspondingly.
As pollution has been shown to be a main vector of SARS-CoV-2 transmission, this ability to identify regions where it may be particularly transmissible and lethal can provide critical advice to implementable public health strategies, such as the ability to monitor, understand, and design NPIs efficiently.
%those responsible for implementing health strategies.

Given that our approach passes several sanity checks by demonstrating that its \pmm concentration predictions reflect ground-truth values/measurements, it would follow that it can make meaningful predictions for locations for which there has never been recordings of the particulate matter (only satellite imagery available).  
Additionally, although we currently do not have rigorous methodology to quantify what makes satellite imagery of different cities `similar', we could address this \eg with dependence tests such as kernel two-sample tests~\cite{GreBorRasSchetal12}. 
Also, data from
multiple satellites could be included to provide greater temporal
resolution, such as from the Sentinel-2 mission~\citep{DRUSCH201225}. With
larger satellite datasets we can expect an increase in the accuracy of our predictions.

\nocite{unet_paper}

\bibliography{bibvid19} 
\bibliographystyle{icml2020}

%\appendix
%\section{Do \emph{not} have an appendix here}
%
%\textbf{\emph{Do not put content after the references.}}
%
%Put anything that you might normally include after the references in a separate
%supplementary file.
%
%We recommend that you build supplementary material in a separate document.
%If you must create one PDF and cut it up, please be careful to use a tool that
%doesn't alter the margins, and that doesn't aggressively rewrite the PDF file.
%pdftk usually works fine. 
%
%\textbf{Please do not use Apple's preview to cut off supplementary material.} In
%previous years it has altered margins, and created headaches at the camera-ready
%stage. 
%%%%%%%%%%%%%%%%%%%%%%%%%%%%%%%%%%%%%%%%%%%%%%%%%%%%%%%%%%%%%%%%%%%%%%%%%%%%%%%
%%%%%%%%%%%%%%%%%%%%%%%%%%%%%%%%%%%%%%%%%%%%%%%%%%%%%%%%%%%%%%%%%%%%%%%%%%%%%%%
%                        END ICML GUIDELINES                                  %
%%%%%%%%%%%%%%%%%%%%%%%%%%%%%%%%%%%%%%%%%%%%%%%%%%%%%%%%%%%%%%%%%%%%%%%%%%%%%%%

\end{document}